\documentclass[conference, letterpaper]{IEEEtran}
\IEEEoverridecommandlockouts    

\usepackage{multirow}
\usepackage{booktabs}
\usepackage{amsmath}
\usepackage{amssymb}
\usepackage{wasysym}
\usepackage[ruled,vlined]{algorithm2e}
\usepackage{graphicx}
\graphicspath{{./figures/}}
\usepackage{etoolbox}
\makeatletter
\patchcmd{\@maketitle}{\newpage}{\newpage\noindent\vspace*{0.25in}}{}{}%
\makeatother
\usepackage[hidelinks,breaklinks]{hyperref}
\usepackage{url}
\usepackage{array}
\usepackage{ragged2e}
\usepackage{siunitx}
\usepackage{stfloats}
\usepackage{cuted}
\usepackage{capt-of}
\usepackage{tikz}
\usepackage{xcolor}
\usepackage{pifont}
\usepackage{enumerate}
\definecolor{matlabBlue}{HTML}{00A7F4}
\definecolor{matlabRed}{HTML}{D20000}

\newcommand{\midcircled}[1]{\tikz[baseline=(char.base)]{
        \node[shape=circle,draw,inner sep=1.2pt, font=\footnotesize] (char) {#1};}}

\newcommand{\sblacksquare}{\scalebox{0.75}{\ensuremath{\blacksquare}}}
\newcommand{\sbigstar}{\scalebox{0.85}{\ensuremath{\bigstar}}}

\title{\LARGE \bf Waypoints Matter: A Systematic Study for Sampling-Based Trajectory Planning}

\author{
	\parbox{\textwidth}{%
		\centering
        Josep M. Barbera$^{*}$, Antonio Artuñedo, Jorge Villagra
        }
	\thanks{The authors are with the AUTOPIA Program at the Centre for Automation and Robotics, CSIC-Universidad Politécnica de Madrid, 28500 Arganda del Rey, Madrid, Spain. $^{*}$Corresponding author: josep.barbera@csic.es.}
}

\IEEEaftertitletext{\vspace{-0.25in}\relax}

\begin{document}
\maketitle
\begin{tikzpicture}[remember picture, overlay]
  \node[anchor=south, yshift=4mm] at (current page.south) {%
    \fbox{%
      \parbox{\dimexpr\textwidth-2\fboxsep-2\fboxrule\relax}{%
        \footnotesize ©2026 IEEE. Personal use of this material is permitted.
        Permission from IEEE must be obtained for all other uses, in any current
        or future media, including reprinting/republishing this material for
        advertising or promotional purposes, creating new collective works, for
        resale or redistribution to servers or lists, or reuse of any copyrighted
        component of this work in other works.%
      }%
    }%
  };
\end{tikzpicture}

\begin{abstract}\sloppy
    Real-time autonomous driving commonly relies on sampling-based trajectory planners that link candidate trajectories to target waypoints along the road centerline. The placement of these waypoints directly impacts both the existence and quality of feasible trajectories. Yet, its effect on planner performance remains largely unexplored. In this paper, we treat waypoint placement as a first-class design variable. We hold the trajectory primitive and candidate budget fixed, and systematically sweep three placement strategies (uniform spacing, an augmented Ramer--Douglas--Peucker variant (RDP*), and a novel curvature-conditioned allocation) across 449 configurations and five CommonRoad maps of increasing geometric complexity. Our results show that the nominal inter-waypoint spacing $d_s$ is the primary performance driver, with large differences in planner reliability attributed to placement alone. Uniform sampling at a well-tuned spacing matches or surpasses both RDP* and the centered curvature variant. The curvature variant offers a small but consistent advantage on geometrically complex roads under reliability-first and balanced weightings, while RDP* never outperforms uniform sampling. These findings suggest that $d_s$ should be treated as the dominant tuning parameter, with geometry-aware strategies reserved for curvature-rich corridors where feasibility is the limiting factor.
\fussy\end{abstract}

\section{Introduction}
\label{sec:introduction}

Real-time trajectory planning is a safety-critical function in autonomous driving. Due to the presence of dynamic obstacles and the need to perform complex maneuvers such as lane changes or overtaking, the vehicle must continuously generate feasible, collision-free paths. Sampling-based planners~\cite{werling2010optimal, gu2013focused, paden2016survey} address this by evaluating candidate trajectories over a local planning horizon, selecting the one that best satisfies kinematic, comfort, and safety constraints. The quality of this candidate set, and ultimately whether any valid trajectory exists at all, depends directly on how the search space is structured. In these planners, it is typically defined by a set of target points (commonly referred to as waypoints) sampled from the road~\cite{piazza2024mptree, lienke2018adhoc}, each serving as a candidate endpoint for trajectory generation. Since generating and validating candidates is computationally expensive~\cite{wursching2021reachable}, the selection of these reference points is not secondary to the choice of trajectory primitive: it determines both the diversity of reachable configurations and the computational efficiency per planning cycle.

In practice, target-point distribution varies across planners, with strategies ranging from fixed-interval grids and line-simplification heuristics to curvature-dependent and data-driven schemes~\cite{autoware2024pathsampler,piazza2024mptree,artunedo2018primitive,artunedo2020ml}. In every case, the placement parameters are configured once for a given planner and scenario rather than systematically varied to assess their effect on planning outcomes.

Waypoint selection has largely been treated as an implementation detail rather than a primary design choice. Yet the few works that have adapted the placement strategy to the driving scenario, whether by constraining intervals through reachable-set bounds~\cite{wursching2021reachable} or by learning evaluation-area parameters~\cite{artunedo2020ml}, report substantial reductions in required samples and computation time while holding the trajectory primitive constant. This suggests that placement exerts a first-order effect on planning performance, motivating a controlled comparison across strategies and parameterizations.

To fill this gap, this paper treats waypoint placement as a first-class design variable and makes the following contributions: 
\begin{enumerate}[(i)]
\item a controlled evaluation framework that fixes the trajectory primitive while systematically varying the placement strategy and its parameters, comparing three strategies over a dense road centerline (equidistant spacing, a Douglas--Peucker simplification augmented to enforce a maximum inter-waypoint distance~\cite{artunedo2019realtime}, and a novel curvature-conditioned allocation) across 449 configurations and five CommonRoad maps~\cite{althoff2017commonroad} of increasing geometric complexity; 
\item  the novel curvature-conditioned allocation itself, which adjusts spacing as an explicit function of local road curvature, providing continuous geometry-aware density control;
\item empirical evidence that the nominal inter-waypoint distance $d_s$ is the dominant performance driver across all strategies and weighting scenarios, together with practical guidelines for strategy and parameter selection.
\end{enumerate}

\section{Related Work} \label{sec:related_work}

\textbf{Waypoint distribution in sampling-based trajectory planners} has been addressed through a variety of strategies depending on the planner architecture. Beyond the most common fixed-interval and simplification-based schemes, uniform grids around a refined corridor are widely used~\cite{li2014,medinalee2022,farag2025efficient}, and some planners adapt placement dynamically based on obstacle geometry or traffic predictions~\cite{homann2019,ogretmen2024}. Data-driven methods go further, learning sampling biases from experience to focus computation on promising regions of the search space~\cite{chaulwar2017,smit2022}. Despite this diversity, the placement rule is in every case part of the planner architecture, tuned for a given scenario and fixed thereafter.

\textbf{Dedicated comparative studies of waypoint placement} remain sparse and most treat it as a secondary concern within broader planner benchmarks. Vilca~et~al.~\cite{vilca2016} formulate multi-criteria waypoint selection for ground robots, jointly optimizing safety, smoothness, and uncertainty over an expanding tree, while Ahmad~et~al.~\cite{ahmad2025} adaptively refine waypoint density around obstacles using piecewise cubic Bézier curves. Most directly related to our work, Artuñedo~et~al.~\cite{artunedo2018primitive} compare equidistant, Douglas--Peucker, and Opheim selection within a large benchmark of path primitives and optimizers, finding that Douglas--Peucker consistently reduces computation time without penalizing path quality. The three selection methods, however, are each tested at a single fixed parameterization and always co-varied with the primitive and optimizer rather than in isolation, so the contribution of placement alone cannot be separated. Our work fixes the trajectory primitive, sweeps each strategy across a wide parameter grid, and evaluates the effect on candidate set quality.

\textbf{Beyond classical trajectory generation}, waypoints appear in several contexts that collectively reinforce the sensitivity of driving performance to their design. In end-to-end models they serve as output representations with tailored loss decompositions~\cite{stelzer2025} or autoregressive generation~\cite{feng2026}, where density and temporal weighting directly affect imitation quality. In reinforcement learning they act as dynamic references for maneuver controllers~\cite{yang2024}, where placement frequency influences policy stability and reactivity. Although the objectives in these works differ from trajectory primitive evaluation, they consistently show that waypoint density and adaptation rules have a measurable impact on planner performance. None of them, however, isolate placement in the controlled, parameter-sweep manner of this paper.

\section{Methodology} \label{sec:methodology}

Autonomous driving in structured environments such as roads or highways typically relies on HD maps. These maps provide the planner with rich information such as lane boundaries, traffic signals, and the road centerline. The latter is particularly important in our case, as we use it as the dense reference curve from which candidate target waypoints are sampled.

\subsection{Waypoints Selection Methods} \label{subsec:waypointselectionmethods}

\begin{figure}[ht]
    \centering
    \includegraphics[width=\linewidth]{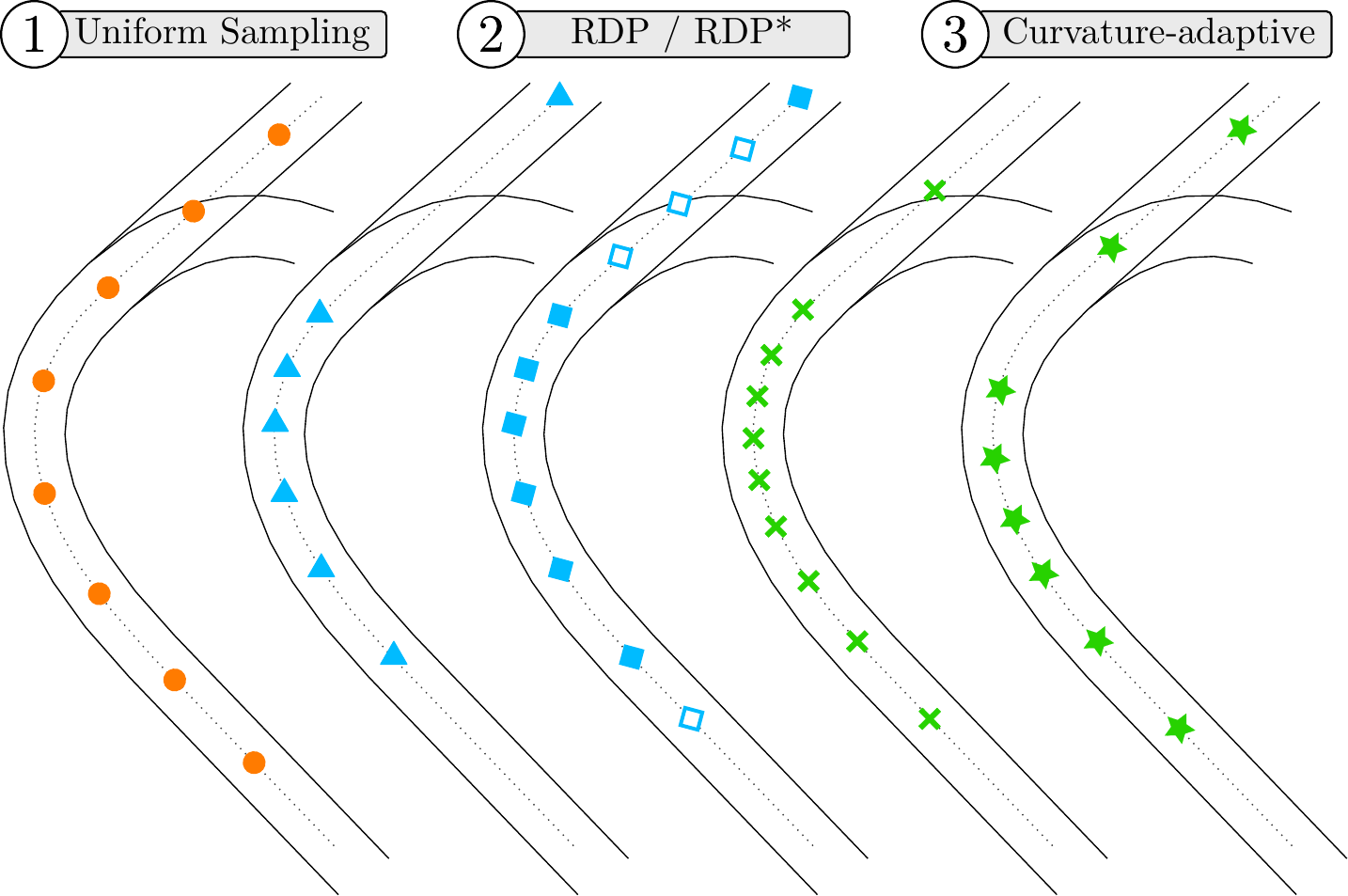}
    \caption{\textbf{Waypoint selection strategies}: \protect\midcircled{1} Uniform sampling, \protect\midcircled{2} RDP/RDP*, and \protect\midcircled{3} Curvature-adaptive sampling. For RDP, $\blacktriangle$ denotes the plain simplification and $\sblacksquare$ the augmented RDP* variant, which enforces a maximum inter-waypoint distance $d_s$ by inserting additional waypoints ($\square$). For curvature-adaptive sampling, \ding{54} uses a centered window while $\sbigstar$ uses a forward window, biasing waypoints upstream of high-curvature segments.}
    \label{fig:waypoint_selection}
\end{figure}

Three waypoint selection strategies are evaluated along the planned corridor centerline $\mathcal{C}$, all parameterized by a nominal arc-length spacing $d_s$, as illustrated in Fig.~\ref{fig:waypoint_selection}.

\paragraph{Uniform sampling} places waypoints at fixed arc-length multiples of $d_s$:

\begin{equation}
    \mathcal{W}_{\text{equi}} = \left\{ \mathbf{p}_i \in \mathcal{C} \mid s_i = i \cdot d_s \right\}.
\end{equation}

\paragraph{RDP*} builds on the classical Ramer--Douglas--Peucker (RDP) algorithm~\cite{douglas1973algorithms}, which simplifies a polyline by iteratively discarding points whose perpendicular deviation from the line segment connecting their neighbors falls below a threshold distance $\varepsilon$. To adapt it for waypoint selection, we first apply RDP to obtain a simplified index set $\mathcal{I}_{\text{RDP}}$, then fill any long segments where the centerline distance $D_{ij}$ between consecutive simplified points exceeds $d_s$. For each such pair $(i,j)$, $n_{ij} = \lfloor D_{ij}/d_s + 0.5 \rfloor$ intermediate indices are generated by linear interpolation between $i$ and $j$, producing a fill set $\mathcal{I}_{\text{fill}}$:
\begin{equation}
    \mathcal{W}_{\text{RDP*}} = \left\{ \mathbf{p}_i \in \mathcal{C} \mid i \in \operatorname{sort}\!\left(\mathcal{I}_{\text{RDP}} \cup \mathcal{I}_{\text{fill}}\right) \right\}.
\end{equation}
In the experiments, we report both the plain RDP simplification (no filling) and the augmented RDP* variant, which isolates the impact of enforcing a maximum inter-waypoint distance on planning performance. We write $\mathcal{W}(\vartheta)$ for the waypoint set produced by whichever method and parameters are encoded in $\vartheta$.

\paragraph{Curvature-adaptive sampling} adjusts the effective spacing according to the local centerline curvature $\kappa(s)$, with a sensitivity parameter $\alpha$. Let $\Delta s_i$ denote the arc-length distance from the most recently placed waypoint to point $i$; a new waypoint is inserted whenever $\Delta s_i$ reaches or exceeds the local spacing threshold:
\begin{align}
    \mathcal{W}_{\text{curv}} & = \left\{ \mathbf{p}_i \in \mathcal{C} \;\middle|\; \Delta s_i \geq d_{\mathrm{local}}(s_i) \right\}, \label{eq:wcurv_def} \\
    d_{\mathrm{local}}(s)     & = \frac{d_s}{1 + \alpha\,|\kappa(s)|}. \label{eq:dlocal_def}
\end{align}
Traversing the discrete centerline sequentially, we accumulate arc length between
consecutive samples and place a waypoint whenever $\Delta s_i$ reaches or exceeds $d_{\mathrm{local}}(s_i)$ at index $i$. Here $\kappa(s)$ is approximated from the discrete curvature values $\kappa_i$ via numerical differentiation of the centerline points and is inherently noisy; a finite smoothing window is applied prior to
evaluating Eq.~\ref{eq:dlocal_def}. We consider both a centered-window variant,
which results in denser sampling around the center of turns, and a forward-window variant that shifts additional waypoints slightly upstream of high-curvature segments. High-curvature regions (large $|\kappa(s)|$) therefore receive denser waypoint coverage, while straight segments revert to a spacing close to the nominal value $d_s$.

\subsection{Systematic Evaluation Framework} \label{subsec:systematicevaluationframework}

\begin{figure*}[t]
    \centering
    \includegraphics[width=\textwidth]{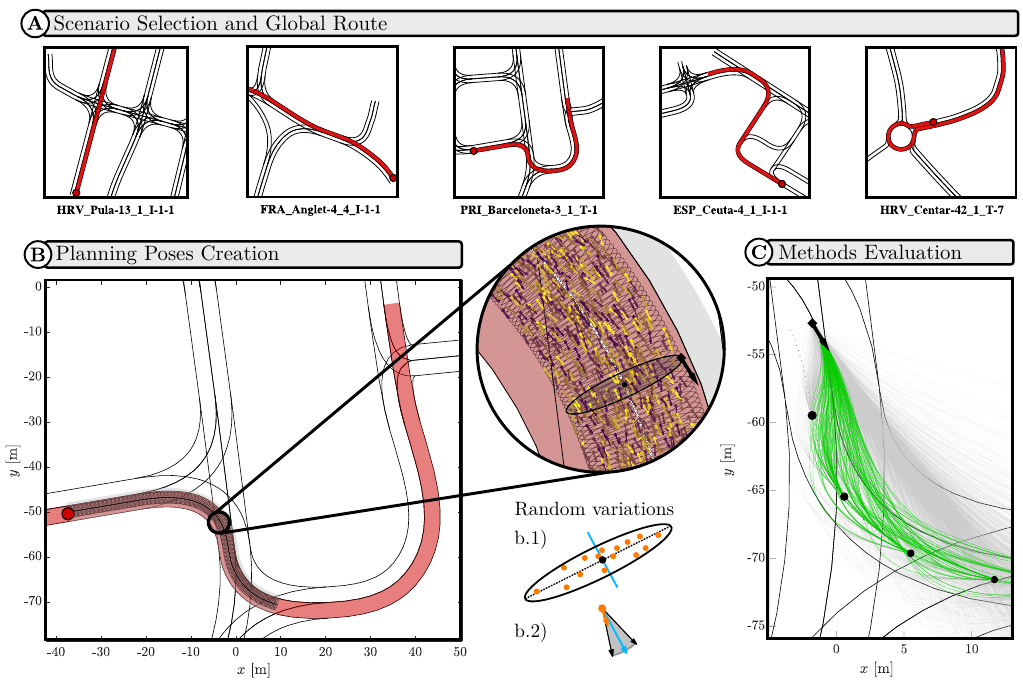}
    \captionof{figure}{\textbf{Systematic evaluation framework.} (A)~\textit{Scenario selection}: five CommonRoad maps span road layouts of increasing geometric complexity. (B)~\textit{Planning poses}: from each map we extract a fixed-length corridor along the centerline and define elliptical sampling regions along it; within each ellipse we draw ego states with uniformly sampled positions~(b.1) and Gaussian perturbations to heading and curvature~(b.2), with curvature magnitude encoded from purple ($|\kappa|=0$) to yellow ($|\kappa|=0.25$). (C)~\textit{Evaluation}: for each ego state, waypoint configuration, and map, the trajectory generator is invoked once, yielding candidate trajectory sets from which we compute validity, length, and diversity metrics.}
    \label{fig:framework}
\end{figure*}

The framework proceeds in three stages, mirroring Fig.~\ref{fig:framework}:
scenario selection~(A), planning pose construction~(B), and systematic
evaluation of each waypoint configuration across all poses and maps~(C).

\vspace{0.33cm} \textbf{(A) Scenario selection.} We evaluate the three waypoint selection strategies within a systematic evaluation framework (see Fig.~\ref{fig:framework}) that keeps the trajectory primitive fixed and varies only the waypoint rule and its parameters. Five CommonRoad maps $\{\mathcal{M}_i\}_{i=1}^{5}$ are selected to span road geometries of increasing complexity, ranging from a straight highway ($\mathcal{M}_1$) to a geometrically demanding urban corridor ($\mathcal{M}_5$), as shown in Fig.~\ref{fig:framework}. For each map, the HD lanelet representation is converted into a driving corridor whose centerline $\mathcal{C}$ is smoothed, shifted laterally so that the ego path lies inside the lane, and trimmed to a fixed planning horizon of $L_h = \SI{60}{m}$ ahead of the origin, yielding a discrete path
\begin{equation}
    \tilde{\mathcal{C}} = \{\mathbf{c}_i\}_{i=1}^{N_c}, \qquad \mathbf{c}_i \in \mathbb{R}^2,
\end{equation}
with $N_c$ discrete points and associated headings $\theta_i$ and curvatures $\kappa_i$.

\vspace{0.33cm} \textbf{(B) Planning poses creation.} \label{subsec:planning_poses_creation} Along $\tilde{\mathcal{C}}$ we construct elliptical sampling regions $E_i$ centered at $\mathbf{c}_i$, aligned with the local tangent, with semi-axes $(b_i, a_i)$ chosen as functions of lane width and lane adjacency (see Fig.~\ref{fig:framework}). For each ellipse, $N_{\text{pts}}$ random ego states are drawn by sampling positions uniformly  over the ellipse area and attaching perturbed headings and curvatures,
\begin{equation}
    x_{i,j} = \big(\mathbf{q}_{i,j}, \theta_{i,j}, \kappa_{i,j}\big),
    \quad j = 1,\dots,N_{\text{pts}},
\end{equation}
where $\mathbf{q}_{i,j}\in E_i$, $\theta_{i,j} = \theta_i + \delta\theta_{i,j}$ with $\delta\theta_{i,j} \sim \mathcal{N}(0, \sigma_\theta^2)$, and $\kappa_{i,j} = \kappa_i + \delta\kappa_{i,j}$ with $\delta\kappa_{i,j} \sim \mathcal{N}(0, \sigma_\kappa^2)$ clipped to $|\kappa_{i,j}| \leq \kappa_{\max}$. Collecting all samples across ellipses yields a finite set of launch states
\begin{equation}
    \mathcal{X} = \{x_{t}\}_{t=1}^{N_T},
\end{equation}
obtained by reindexing the pairs $(i,j)$. The resulting set sizes are $N_T=9045$ for $\mathcal{M}_1$ and $\mathcal{M}_2$, $N_T=8985$ for $\mathcal{M}_3$, and $N_T=9015$ for $\mathcal{M}_4$ and $\mathcal{M}_5$.

\vspace{0.33cm} \textbf{(C) Methods evaluation.} Each selection strategy is evaluated across a range of parameter configurations $\vartheta \in \Theta$, where $\Theta$ spans combinations of nominal spacing $d_s$, RDP tolerance $\varepsilon$, curvature sensitivity $\alpha$, and smoothing mode $\gamma$ (forward vs.\ centered),
\begin{equation}
    \Theta = \big\{ (d_s, \varepsilon, \alpha, \gamma) \big\},
\end{equation}
with a total of $449$ configurations across all methods (see Table~\ref{tab:scan_params}). For every combination of map $\mathcal{M}_i$, configuration $\vartheta$, and launch state $x \in \mathcal{X}$, the trajectory generator produces a candidate set using $\mathcal{W}(\vartheta)$ along $\tilde{\mathcal{C}}$ with a fixed budget of $N_{\text{cand}} = 4000$ candidates distributed uniformly over waypoints. The trajectory primitive is a degree-5 Bézier curve with G2-continuous boundary conditions at both the ego state and the target waypoint~\cite{artunedo2019realtime}. Four scalar metrics are extracted to characterize each candidate set and aggregated into a weighted score.

\begin{table}[!t]
    \centering
    \footnotesize
    \setlength{\tabcolsep}{5pt}
    \caption{Parameter ranges for the waypoint-selection study.}
    \label{tab:scan_params}
    \begin{tabular}{c c c c c c r}
    \hline
        Batch & Method & $d_s$ [\si{m}] & $\varepsilon$ [\si{m}] & $\alpha$ [\si{m}] & $\gamma$ & $N$ \\
        \hline
        1 & $\bullet$            & $0.5{:}0.25{:}10$ & --                   & --             & -- & 39  \\
        2 & $\blacktriangle$         & --                & $0.01{:}0.01{:}0.10$ & --             & -- & 10  \\
        3 & $\blacktriangle$         & --                & $0.15{:}0.05{:}1.00$ & --             & -- & 18  \\
        4 & $\sblacksquare$          & $\{2,3,5,7\}$     & $0.01{:}0.01{:}0.10$ & --             & -- & 40  \\
        5 & $\sblacksquare$          & $\{2,3,5,7\}$     & $0.15{:}0.05{:}1.00$ & --             & -- & 72  \\
        6 & \ding{54}      & $2{:}1{:}10$      & --                   & $5{:}2.5{:}40$ & 1  & 135 \\
        7 & $\sbigstar$     & $2{:}1{:}10$      & --                   & $5{:}2.5{:}40$ & 0  & 135 \\
        \hline
        \multicolumn{6}{l}{Total} & 449 \\
        \hline
        \end{tabular}
    \begin{minipage}{\linewidth}
        \vspace{0.15cm}
        \scriptsize
        \textit{Method symbols:} $\bullet$: uniform sampling; $\blacktriangle$: plain RDP; $\sblacksquare$: RDP* (with $d_s$ gap-filling); \ding{54}: curvature-adaptive centered window ($\gamma = 1$); $\sbigstar$: curvature-adaptive forward window ($\gamma = 0$). Batch~1 yields 39 configurations; batches~2--3 and 4--5 together yield 140; batches~6--7 yield 270; totalling 449.
    \end{minipage}
\end{table}

\textbf{Metrics.} Let $\mathcal{V}$ denote the valid candidates with lengths $\{\ell_k\}_{k\in\mathcal{V}}$ and $n_v = |\mathcal{V}|$ obtained for each planning request. Reliability is captured by a per-map success rate. Let $n_{\text{fail}}$ be the number of launch states returning zero valid candidates; then
\begin{equation}
    f_1 = 1 - \frac{n_{\text{fail}}}{N_T},
\end{equation}
where $N_T$ is the total number of launch states for that map (see Section~\ref{subsec:planning_poses_creation}). Three additional metrics characterize each candidate set:
\begin{align}
    f_2 & = \frac{n_v}{N_{\text{cand}}},                                                        \\
    f_3 & = \frac{\bar{L}}{L_h}, \quad \bar{L} = \frac{1}{n_v} \sum_{k \in \mathcal{V}} \ell_k, \\
    f_4 & = -\frac{1}{\log B} \sum_{b=1}^{B} p_b \log p_b,
\end{align}
where $f_2$ is the fraction of valid candidates out of the generation budget $N_{\text{cand}}$, $f_3$ the mean trajectory length normalized by $L_h$, and $f_4$ the normalized Shannon entropy over $B=3$ equal-width length bins with $p_b$ the fraction of valid trajectories whose length falls in bin $b$. All four metrics are higher-is-better and lie in $[0,1]$. Results are summarized via the weighted score
\begin{equation}
    J = w_1\,f_1 + w_2\,f_2 + w_3\,f_3 + w_4\,f_4,
\end{equation}
where the weights sum to one and reflect a priority ranking of the four objectives. We propose $w_1 = 0.70$, $w_2 = 0.20$, $w_3 = 0.08$, and $w_4 = 0.02$ as a balanced option with priority to metrics $f_1$ and $f_2$. Reliability ($f_1$, 70\%) receives the largest weight because a configuration that produces no valid trajectory in a significant fraction of situations is unsafe regardless of candidate quality. The candidate count ($f_2$, 20\%) ranks second, as a larger set of valid options improves downstream trajectory selection for the current driving context. Trajectory length ($f_3$, 8\%) ranks third, since longer trajectories extend the replanning horizon and reduce the frequency of candidate-generation cycles. Diversity ($f_4$, 2\%) is weighted lowest: longer trajectories inherently reduce candidate spread, so $f_4$ and $f_3$ are partly antagonistic, and the combined weight allocated to both is necessarily small relative to the dominant terms $f_1$ and $f_2$.

\textbf{Implementation.} The trajectory generator is implemented in C++ using the Lanelet2 library for map parsing \cite{poggenhans2018lanelet2}. Experiment orchestration uses MATLAB and Python scripts that dispatch each configuration as an independent process, parallelised across 32 cores of an Intel Core i9-14900KS (64\,GiB RAM). Each single-configuration run takes approximately \SI{330}{s}; the full experiment of 449 configurations across five maps totals approximately 8\,days and 14\,hours of wall-clock time. Fixed parameters common to all configurations are listed in Table~\ref{tab:params}.

\begin{table}[h]
    \centering
    \caption{Fixed experimental parameters common to all configurations.}
    \label{tab:params}
    \begin{tabular}{lcc}
        \toprule
        \textbf{Parameter} & \textbf{Symbol} & \textbf{Value} \\
        \midrule
        Planning horizon        & $L_h$                  & \SI{60}{m}           \\
        Candidate budget        & $N_{\text{cand}}$      & 4000                 \\
        Points per ellipse      & $N_{\text{pts}}$       & 15                   \\
        Base ellipse semi-axes       & $(a,\, b)$             & (\SI{1.30}{m},\ \SI{0.25}{m}) \\
        Heading noise $\sigma$  & $\sigma_{\theta}$      & \SI{10}{\degree}     \\
        Curvature noise $\sigma$& $\sigma_{\kappa}$      & \SI{0.1}{m^{-1}}     \\
        Max curvature           & $\kappa_{\max}$        & \SI{0.25}{m^{-1}}    \\
        \bottomrule
    \end{tabular}
\end{table}

\section{Results} \label{sec:results}

Uniform sampling ($\bullet$) serves as the reference baseline; RDP* ($\sblacksquare$) and the curvature-adaptive strategies (\ding{54}, $\sbigstar$) are evaluated against it across all metrics and parameter settings. Results are reported by aggregating each metric across the five maps $\{\mathcal{M}_i\}_{i=1}^{5}$. Since each map represents a distinct road geometry, aggregation yields a single value that reflects general behaviour rather than performance on any specific scenario. The failure count $\sum n_{\text{fail}}$ is summed across maps, as it is a raw count whose total reflects the overall reliability burden. The scalar metrics $f_1$ through $f_4$ and the weighted score $J$ are averaged across maps. For the remainder of the paper, $f_1$ through $f_4$ and $J$ refer to these cross-map aggregates unless stated otherwise. For interpretability, $f_2$ and $f_3$ are denormalized in the figures as $\bar{n}_{\text{valid}} = f_2 \cdot N_{\text{cand}}$ (as a candidate count) and $\bar{L} = f_3 \cdot L_h$ (in metres). Together, these aggregated values are used to rank configurations and identify the best-performing parameter settings across all evaluated road geometries.

\subsection{Uniform Sampling} \label{subsec:uniform_sampling}
\begin{figure}[t]
    \centering
    \includegraphics[width=\linewidth]{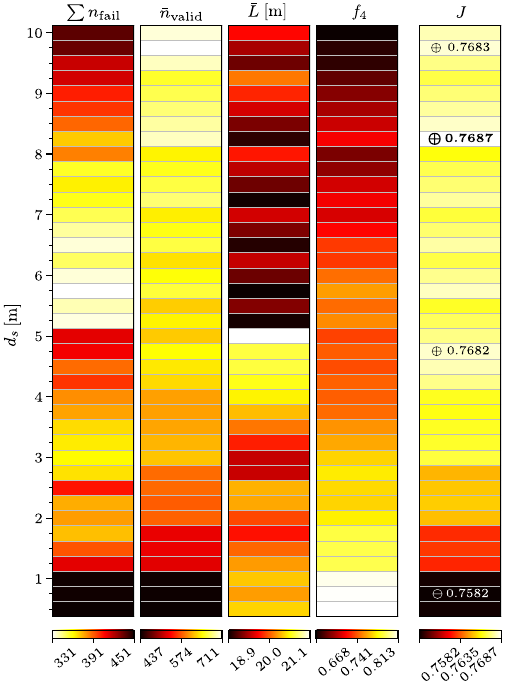}
    \caption{\textbf{Uniform sampling performance across maps}. The waypoint separation distance $d_s$ is varied from \SI{0.5}{m} to \SI{10}{m} in increments of \SI{0.25}{m}. For each variation, columns report (left to right): total failed planning queries, mean valid candidates, mean trajectory length, and mean diversity; aligned with $f_1$--$f_4$, these contribute 70\%, 20\%, 8\%, and 2\% respectively to the weighted score $J$ shown in the last column. Brighter cells indicate more desirable values. The bold entry marks the best $J$; $\boldsymbol{\oplus}$ indicates the top-3 configurations and $\boldsymbol{\ominus}$ the minimum. A moderate spacing around $d_s \approx \SI{8}{m}$ achieves the best balance between reliability and candidate yield.}
    \label{fig:method0_heatmap_metrics}
\end{figure}

For uniform sampling, results across 39 values of $d_s$ are shown in Fig.~\ref{fig:method0_heatmap_metrics}. In terms of reliability, an average of 391 out of approximately 9\,000 planning queries fail across the five maps, with the minimum of 331 failures achieved at $d_s = \SI{5.75}{m}$. The mean candidate count grows steadily with $d_s$, ranging from 400 to 711 and peaking at $d_s = \SI{9.75}{m}$. Mean trajectory length varies narrowly between 19 and 21\,m, with $d_s = \SI{5}{m}$ maximizing it at $\bar{L} = \SI{21.1}{m}$. Diversity shows more variation (0.67--0.81): smaller spacing produces more spread in the candidate lengths, while larger spacing tends to cluster trajectories around similar lengths. Under the weighted score $J$, the best configuration is $d_s = \SI{8.25}{m}$, followed by $d_s = \SI{9.75}{m}$ and $d_s = \SI{4.75}{m}$.

\subsection{RDP/RDP*} \label{subsec:rdp_star}
\begin{figure}[t]
    \centering
    \includegraphics[width=\linewidth]{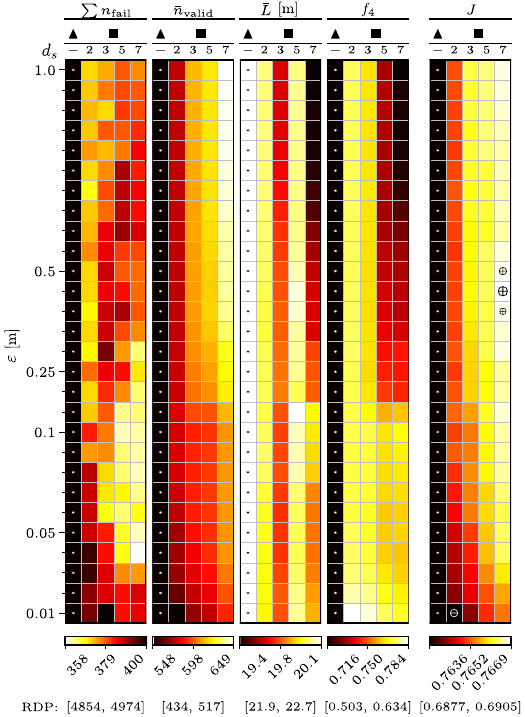}
    \caption{\textbf{RDP and RDP* performance across maps}. Plain RDP ($\blacktriangle$) and the augmented RDP* variant ($\sblacksquare$) are evaluated as a function of simplification tolerance $\varepsilon$, swept from 0.01 to 1.00\,m; for RDP*, gap-filling is applied at $d_s \in \{2, 3, 5, 7\}$\,m. Columns and color convention follow Fig.~\ref{fig:method0_heatmap_metrics}. Color scales are set independently for RDP and RDP* due to their markedly different metric ranges. RDP* consistently outperforms plain RDP across all metrics except trajectory length; within RDP*, the best configurations favor large $d_s$ and moderate to high $\varepsilon$, with the top three at $d_s = 7$\,m and $\varepsilon \in \{0.35, 0.45, 0.50\}$\,m. Annotation follow the convention of Fig.~\ref{fig:method0_heatmap_metrics}: bold for best $J$, $\boldsymbol{\oplus}$ for top-3, $\boldsymbol{\ominus}$ for the minimum.}
    \label{fig:method1_heatmap}
\end{figure}

For RDP and RDP*, the simplification tolerance $\varepsilon$ is swept at two resolutions: finely from 0.01 to 0.10\,m (step 0.01\,m) and more sparsely from 0.15 to 1.00\,m (step 0.05\,m). This change in resolution is visible in Fig.~\ref{fig:method1_heatmap}, particularly in the $\bar{n}_{\text{valid}}$ and $f_4$ columns, where the gradient shifts noticeably above $\varepsilon = 0.15$\,m. In terms of metric ranges, plain RDP consistently produces worse results than RDP* across all metrics except average trajectory length; to preserve readability, RDP values are excluded from the RDP* color scale in the figure and saturated to the extremes of the RDP* range so they don't affect the color gradient.

Focusing on RDP*, reliability is best in the region $\varepsilon \in [0.05, 0.25]$\,m for $d_s \in \{5, 7\}$\,m, with the minimum of 358 failed queries at $\varepsilon = 0.04$\,m and $d_s = 7$\,m. The mean candidate count grows toward larger $\varepsilon$ and $d_s$, reaching a maximum at 649 valid candidates for $(\varepsilon, d_s) = (\SI{1.00}{m},\, \SI{7}{m})$. Mean trajectory length is highest for $d_s \in \{2, 3\}$\,m, reaching its maximum at $d_s = 5$\,m and $\varepsilon = 0.15$\,m, while larger $d_s$ values with high $\varepsilon$ produce shorter trajectories. Diversity follows the opposite trend from candidate count: smaller $\varepsilon$ and smaller $d_s$ retain more waypoints and produce more varied trajectory sets, with the highest $f_4$ at $\varepsilon = 0.01$\,m and $d_s = 2$\,m.

The weighted score $J$ is dominated by $d_s$, with higher values generally preferred, and shows a secondary improvement with increasing $\varepsilon$. The three best configurations all sit at $d_s = 7$\,m with $\varepsilon \in \{0.45, 0.50, 0.35\}$\,m ranked in order. Even at its best, RDP* does not exceed the top uniform configurations at comparable $d_s$.

\subsection{Curvature-Adaptive Sampling} \label{subsec:curvature_adaptive_sampling}
\begin{figure*}[t]
    \centering
    \includegraphics[width=\textwidth]{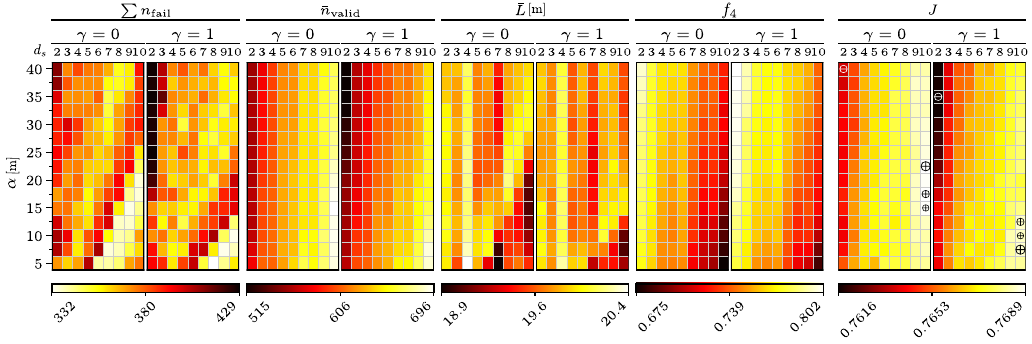}
    \captionof{figure}{\textbf{Curvature-adaptive sampling performance across maps}. For this method, the sensitivity parameter $\alpha$ (how much curvature affects the waypoint spacing) is varied from 5 to 40 in increments of 2.5, while $d_s$ spans from \SI{2}{m} to \SI{10}{m} in increments of \SI{1}{m}. Color scales are shared between the two smoothing modes, centered-window ($\gamma = 1$) and forward-window ($\gamma = 0$), for direct comparison. The forward window consistently shifts performance toward better reliability and candidate yield; the top configurations under $J$ are at $\gamma = 0$, $d_s = 10$\,m, and moderate $\alpha \in \{15, 17.5, 22.5\}$\,m. Annotation convention as in Fig.~\ref{fig:method0_heatmap_metrics}.}
    \label{fig:method2_heatmap}
\end{figure*}

For curvature-adaptive sampling, the sensitivity parameter $\alpha$ is swept from 5 to 40\,m in steps of 2.5\,m, combined with integer nominal spacings $d_s \in [2, 10]$\,m, and evaluated for both smoothing modes: centered window ($\gamma = 1$) and forward window ($\gamma = 0$).

The most notable pattern in Fig.~\ref{fig:method2_heatmap} is that the results for $\gamma = 0$ closely mirror those for $\gamma = 1$ shifted toward smaller $d_s$. This occurs because the forward window anticipates curvature earlier along the path, effectively acting as if the planner were operating with denser spacing. As a result, the low-$d_s$ region where $\gamma = 1$ performs poorly largely disappears from the $\gamma = 0$ evaluations. For both $\sum n_{\text{fail}}$ and $\bar{n}_{\text{valid}}$, the dominant trend is that smaller $d_s$ and larger $\alpha$ produce worse outcomes, with $\bar{n}_{\text{valid}}$ showing a smoother gradient than $\sum n_{\text{fail}}$. The diversity metric $f_4$ and mean length $\bar{L}$ exhibit the reverse behaviour: smaller $d_s$ and higher $\alpha$ favor diversity, while $\bar{L}$ is less sensitive to $\alpha$. Since $J$ heavily weights reliability and candidate count, the overall ranking follows the same trends as those two metrics.

The best individual configurations are $(\gamma, d_s, \alpha) = (1, 8, 5)$ for fewest failures, $(0, 10, 15)$ for mean candidate count, $(0, 4, 5)$ for trajectory length, and $(1, 2, 40)$ for diversity. Under the weighted score $J$, the top three configurations are all at $\gamma = 0$ and $d_s = 10$\,m, with $\alpha \in \{22.5, 17.5, 15.0\}$\,m.

\section{Discussion} \label{sec:discussion}

Under every weighting scenario in Table~\ref{tab:best_configs}, the top-ranked configurations share one characteristic: large nominal spacing ($d_s$). Curvature-adaptive forward sampling ($\sbigstar$) and uniform sampling ($\bullet$) both reach the top positions under balanced weights, separated by a negligible margin. Plain RDP ($\blacktriangle$) falls behind on the highway map ($\mathcal{M}_1$), where a single distant waypoint combined with random heading choices causes most candidates to leave the lane. RDP* ($\sblacksquare$) corrects this through gap-filling but its metrics span a narrower range that falls inside those of the two leading strategies, leaving its best configuration scoring lower. The difference between $\sbigstar$ and $\bullet$ is that curvature-adaptive sampling keeps all four metrics simultaneously favorable, which is what $J$ rewards. In all cases, \textit{choosing the right spacing matters more than choosing the right strategy}.

\begin{table*}[t]
    \centering
    \caption{Top-10 configurations ranked by $J$ for three weight scenarios.}
    \label{tab:best_configs}
    \renewcommand{\arraystretch}{1.1}
\begin{footnotesize}
\setlength{\tabcolsep}{3.5pt}
\begin{tabular*}{0.982\linewidth}{@{\extracolsep{\fill}} l  c  c  S[table-format=2.2]  S[table-format=1.2]  S[table-format=2.1]  c  S[table-format=5.0]  S[table-format=4.1]  S[table-format=2.2]  S[table-format=1.4]  S[table-format=1.6]}
    \toprule
    \multirow{2}{*}{\shortstack[c]{Weights\\\scriptsize$(w_1,w_2,w_3,w_4)$ [\%]}} & \multirow{2}{*}{\#} & \multirow{2}{*}{Method} & \multicolumn{4}{c}{Parameters} & \multicolumn{5}{c}{Metrics} \\
    \cmidrule(lr){4-7} \cmidrule(lr){8-12}
    & & & {$d_s$} & {$\varepsilon$} & {$\alpha$} & {$\gamma$} & {$\sum n_{\mathrm{fail}}$} & {$\bar{n}_{\mathrm{valid}}$} & {$\bar{L}$ [m]} & {$f_4$} & {$J$} \\
    \midrule
    \multirow{10}{*}{\shortstack[c]{\textbf{Balanced}\\\smallskip$(70,20,8,2)$[\%]}}
        & 1  & \makebox[1em][c]{$\bigstar$} & {$10$} & {$-$} & {$22.5$} & {$0$} & {348} & {683.9} & {19.40} & {0.7137} & {\textbf{0.768940}} \\
        & 2  & \makebox[1em][c]{$\bigstar$} & {$10$} & {$-$} & {$17.5$} & {$0$} & {\textbf{336}} & {693.2} & {19.06} & {0.7022} & {0.768912} \\
        & 3  & \makebox[1em][c]{$\bigstar$} & {$10$} & {$-$} & {$15.0$} & {$0$} & {343} & {\textbf{696.4}} & {19.06} & {0.6986} & {0.768885} \\
        & 4  & \makebox[1em][c]{$\bigstar$} & {$10$} & {$-$} & {$20.0$} & {$0$} & {\textbf{336}} & {687.9} & {19.12} & {0.7098} & {0.768878} \\
        & 5  & \makebox[1em][c]{$\bullet$} & {$8.25$} & {$-$} & {$-$} & {$-$} & {371} & {694.0} & {19.06} & {0.7194} & {0.768748} \\
        & 6  & \makebox[1em][c]{$\bigstar$} & {$10$} & {$-$} & {$25.0$} & {$0$} & {360} & {673.8} & {19.71} & {0.7143} & {0.768678} \\
        & 7  & \makebox[1em][c]{$\bigstar$} & {$10$} & {$-$} & {$27.5$} & {$0$} & {379} & {673.1} & {19.87} & {0.7197} & {0.768666} \\
        & 8  & \makebox[1em][c]{$\bigstar$} & {$10$} & {$-$} & {$30.0$} & {$0$} & {390} & {670.3} & {\textbf{20.02}} & {0.7216} & {0.768594} \\
        & 9  & \makebox[1em][c]{\ding{54}} & {$10$} & {$-$} & {$7.5$} & {$1$} & {339} & {690.0} & {19.05} & {0.6957} & {0.768567} \\
        & 10 & \makebox[1em][c]{$\bigstar$} & {$9$} & {$-$} & {$37.5$} & {$0$} & {359} & {656.1} & {19.87} & {\textbf{0.7399}} & {0.768533} \\
    \midrule
    \multirow{10}{*}{\shortstack[c]{\textbf{Reliability-first}\\\smallskip$(90,8,1,1)$[\%]}}
        & 1  & \makebox[1em][c]{$\bigstar$} & {$10$} & {$-$} & {$17.5$} & {$0$} & {336} & {693.2} & {19.06} & {0.7022} & {\textbf{0.917358}} \\
        & 2  & \makebox[1em][c]{$\bigstar$} & {$10$} & {$-$} & {$20.0$} & {$0$} & {336} & {687.9} & {19.12} & {0.7098} & {0.917338} \\
        & 3  & \makebox[1em][c]{$\bigstar$} & {$10$} & {$-$} & {$15.0$} & {$0$} & {343} & {\textbf{696.4}} & {19.06} & {0.6986} & {0.917246} \\
        & 4  & \makebox[1em][c]{$\bullet$} & {$5.75$} & {$-$} & {$-$} & {$-$} & {\textbf{331}} & {657.3} & {18.94} & {\textbf{0.7548}} & {0.917244} \\
        & 5  & \makebox[1em][c]{\ding{54}} & {$10$} & {$-$} & {$10.0$} & {$1$} & {334} & {680.6} & {19.10} & {0.7085} & {0.917217} \\
        & 6  & \makebox[1em][c]{\ding{54}} & {$10$} & {$-$} & {$7.5$} & {$1$} & {339} & {690.0} & {19.05} & {0.6957} & {0.917169} \\
        & 7  & \makebox[1em][c]{$\bigstar$} & {$10$} & {$-$} & {$22.5$} & {$0$} & {348} & {683.9} & {19.40} & {0.7137} & {0.917104} \\
        & 8  & \makebox[1em][c]{$\bigstar$} & {$9$} & {$-$} & {$15.0$} & {$0$} & {335} & {664.7} & {\textbf{19.45}} & {0.7231} & {0.917082} \\
        & 9  & \makebox[1em][c]{\ding{54}} & {$9$} & {$-$} & {$7.5$} & {$1$} & {334} & {661.3} & {19.41} & {0.7211} & {0.917008} \\
        & 10 & \makebox[1em][c]{$\bullet$} & {$6.50$} & {$-$} & {$-$} & {$-$} & {336} & {658.9} & {19.03} & {0.7337} & {0.916982} \\
    \midrule
    \multirow{10}{*}{\shortstack[c]{\textbf{Coverage-first}\\\smallskip$(8,90,1,1)$[\%]}}
        & 1  & \makebox[1em][c]{$\bullet$} & {$9.75$} & {$-$} & {$-$} & {$-$} & {434} & {\textbf{711.3}} & {19.46} & {0.6751} & {\textbf{0.249317}} \\
        & 2  & \makebox[1em][c]{$\bullet$} & {$10.00$} & {$-$} & {$-$} & {$-$} & {436} & {699.9} & {\textbf{19.76}} & {0.6679} & {0.246725} \\
        & 3  & \makebox[1em][c]{$\bigstar$} & {$10$} & {$-$} & {$15.0$} & {$0$} & {343} & {696.4} & {19.06} & {0.6986} & {0.246291} \\
        & 4  & \makebox[1em][c]{$\bullet$} & {$8.25$} & {$-$} & {$-$} & {$-$} & {371} & {694.0} & {19.06} & {\textbf{0.7194}} & {0.245926} \\
        & 5  & \makebox[1em][c]{$\bigstar$} & {$10$} & {$-$} & {$17.5$} & {$0$} & {\textbf{336}} & {693.2} & {19.06} & {0.7022} & {0.245613} \\
        & 6  & \makebox[1em][c]{$\bullet$} & {$9.50$} & {$-$} & {$-$} & {$-$} & {417} & {694.1} & {19.27} & {0.6761} & {0.245462} \\
        & 7  & \makebox[1em][c]{$\bigstar$} & {$10$} & {$-$} & {$10.0$} & {$0$} & {371} & {692.8} & {19.21} & {0.6891} & {0.245376} \\
        & 8  & \makebox[1em][c]{$\bigstar$} & {$10$} & {$-$} & {$12.5$} & {$0$} & {350} & {691.7} & {19.12} & {0.6906} & {0.245154} \\
        & 9  & \makebox[1em][c]{$\bigstar$} & {$10$} & {$-$} & {$7.5$} & {$0$} & {362} & {691.6} & {19.33} & {0.6831} & {0.245074} \\
        & 10 & \makebox[1em][c]{\ding{54}} & {$10$} & {$-$} & {$7.5$} & {$1$} & {339} & {690.0} & {19.05} & {0.6957} & {0.244840} \\
    \bottomrule
\end{tabular*}
\end{footnotesize}

    \begin{minipage}{\linewidth}
        \vspace{0.15cm}
        \scriptsize
        Across all three weighting scenarios, large nominal spacing $d_s$ consistently dominates the top ranks. Curvature-adaptive forward sampling ($\sbigstar$) and uniform sampling ($\bullet$) trade places near the top when reliability carries significant weight, while the centered variant (\ding{54}) emerges mainly when coverage is prioritized. Bold values mark the best entry in each column.
    \end{minipage}
\end{table*}

The small but consistent advantage of curvature-adaptive forward sampling ($\sbigstar$) over uniform spacing ($\bullet$) reveals where geometry-aware placement adds value. As Fig.~\ref{fig:method2_heatmap} shows, the forward-window ($\gamma = 0$) heatmap closely mirrors the centered-window ($\gamma = 1$) heatmap shifted toward lower $d_s$, meaning the \textit{forward window pre-densifies the road segment the planner is about to enter}. Candidate trajectories toward an upcoming curve therefore start from a denser target set, compensating for the higher kinematic rejection rate that curves impose. The centered window also reaches competitive scores, but its density peak aligns with the midpoint of the turn rather than its entry, arriving too late to recover candidates lost to curvature-induced infeasibility. RDP* ($\sblacksquare$) fails to compete because at high simplification tolerance, the algorithm discards most centerline points and the gap-filling step reinserts them at uniform intervals, converging to uniform sampling; at low tolerance, it retains too many points and fragments the candidate budget, mirroring the effect of small $d_s$ in uniform sampling. Prior work~\cite{artunedo2018primitive} reported benefits for RDP when placement was co-varied with the trajectory primitive, but the isolated comparison here shows no such advantage.

The four metrics split into two groups that respond to spacing in opposite ways. Reliability ($f_1$) and candidate count ($f_2$) both improve with larger $d_s$. Since $N_{\text{cand}}$ is fixed and split uniformly across waypoints, wider spacing assigns more candidates per target, raising the probability that at least one is feasible. Diversity ($f_4$) and, to a lesser extent, mean trajectory length ($f_3$) favor smaller $d_s$, where denser waypoints produce more varied candidate sets. The three weighting scenarios in Table~\ref{tab:best_configs} illustrate this directly. Under reliability-dominant and balanced weights, curvature-adaptive forward ($\sbigstar$) and uniform ($\bullet$) interleave at the top with large $d_s$; under coverage-dominant weights, uniform at the largest tested spacing rises first. The rankings are therefore robust to any weighting that prioritizes reliability, but shift when diversity is the priority, in which case smaller $d_s$ combined with curvature adaptation becomes preferable. As a practical guideline, without curvature information, uniform spacing in the upper tested range requires no tuning and is near-competitive with any more complex scheme; with curvature information, a forward-window adaptive rule at large nominal spacing and moderate $\alpha$ extracts a small additional margin, particularly on geometrically complex roads.

\section{Conclusions} \label{sec:conclusion}

This paper treated waypoint placement as a first-class design variable and evaluated three selection strategies across 449 configurations and five CommonRoad maps. The results confirm that under a fixed trajectory primitive and candidate budget, varying placement alone produces failure-rate swings from 331 to well over 400. The nominal inter-waypoint distance $d_s$ is the primary performance driver. Curvature-adaptive forward sampling provides a small but consistent advantage over uniform spacing under balanced and reliability-first weightings (Table~\ref{tab:best_configs}). RDP* never outperforms the simpler alternatives under any weighting scenario.

The candidate budget $N_{\text{cand}}$ is distributed uniformly over waypoints; a non-uniform allocation could decouple waypoint density from per-target sampling effort. The evaluation uses static maps without dynamic obstacles, so the interaction between waypoint density and collision-avoidance feasibility remains open. The weighted score $J$ relies on manually selected weights, and replacing it with explicit Pareto-front extraction would remove the dependence on a single scalar ranking. Finally, a single degree-5 Bézier primitive underlies all experiments; whether the spacing-dominance finding transfers to other primitives such as polynomial spirals or clothoids remains to be determined. Understanding how waypoint placement interacts with candidate allocation, trajectory generation, and obstacle avoidance is the natural next step toward joint optimization of these interdependent design choices.

\section*{Acknowledgment}

This work was supported in part by the Chips Joint Undertaking through project SHAPEFUTURE under Grant 101139996, in part by MCIN/AEI/10.13039/501100011033 with the project SHAPEFUTURE (PCI2024-153527) and in part by the Community of Madrid through the SEGVAUTO5G-CM Programme (TEC-2024/ECO-277).

\begin{sloppypar}
\bibliographystyle{IEEEtran}
\bibliography{root}
\end{sloppypar}
\end{document}